\documentclass[
  journal=medium,
  manuscript=technical-report,
  year=2026,
  volume=0,
]{cup-journal}
\usepackage[nopatch]{microtype}

\usepackage[T1]{fontenc}    % use 8-bit T1 fonts
\usepackage{hyperref}       % hyperlinks
\usepackage{tikz}          % inline TikZ figure sources
\usetikzlibrary{positioning,arrows.meta,fit,backgrounds,calc}

\usepackage{subcaption}
\usepackage{listings}
\usepackage{tcolorbox}
\tcbuselibrary{skins,breakable,listings}

\colorlet{tbrowcolor}{white}
\colorlet{tbheadcolor}{white}

\newcommand{\nxmndr}{\texttt{nxmndr}}
\newcommand{\nxmndrqgis}{\texttt{nxmndr-qgis}}

\title{Anaximander: Interactively Running Geospatial \\ Deep Learning Models on Any Compute Backend}

\author{Satej S. Soman\textsuperscript{\textdagger}}
\affiliation{Microsoft AI for Good Research Lab}

\author{Akram Zaytar}
\affiliation{Microsoft AI for Good Research Lab}
\email[Author2]{akramzaytar@microsoft.com}

\author{Girmaw A. Tadesse}
\affiliation{Microsoft AI for Good Research Lab}

\author{Gilles Q. Hacheme}
\affiliation{Microsoft AI for Good Research Lab}

\author{Muhammad S. Danish}
\affiliation{Microsoft AI for Good Research Lab}

\author{Inbal Becker-Reshef}
\affiliation{Microsoft AI for Good Research Lab}

\author{Rahul Dodhia}
\affiliation{Microsoft AI for Good Research Lab}

\author{Juan Lavista Ferres}
\affiliation{Microsoft AI for Good Research Lab}

\keywords{QGIS, Remote Sensing, Deep Learning, Computer Vision}

\makeatletter
\patchcmd{\@maketitle}
  {{\volumefont\cup@vol}, }
  {}
  {}{}
\patchcmd{\ps@plain}
  {\@author@i}
  {{\let\textsuperscript\@gobble\@author@i}}
  {}{}
\renewcommand\paragraph{\@startsection{paragraph}{4}{\z@}%
  {3.25ex \@plus1ex \@minus.2ex}%
  {-1em}%
  {\sffamily\bfseries}}
\pretocmd{\cup@contact@details}%
  {{\textsuperscript{\textdagger}Also School of Information, UC Berkeley. Work
  done while the author was an AI Resident at the Microsoft AI for Good Research Lab,
  Nairobi, Kenya.}\par}%
  {}{}
\makeatother

\begin{document}
\maketitle

\begin{abstract}
Applying deep learning models to satellite imagery from within geographic information systems (GIS) remains high-friction for remote sensing practitioners. Models arrive in incompatible formats and target different compute environments, from local workstations to serverless cloud services. As a result, every evaluation demands custom deployment, tiling, and georeferencing code before a single prediction reaches the analyst's map. This friction discourages systematic comparison in a domain where model choice directly affects operational outcomes such as field delineation, crop monitoring, and disaster response. We present Anaximander, an open-source system that unifies model source and compute location choice behind one interactive interface. The system's backend is an inference server that loads models from multiple commonly used sources and serves them on any accessible compute backend. The server provides session management and model caching, and streams results back per tile. The backend is paired with a QGIS plugin that drives tiling, result reassembly, georeferencing, and real-time per-tile status visualization. An additional user-interface path injects layer legends as prompts into vision-language models. We demonstrate the system in a code-free side-by-side comparison of three heterogeneous models on an agricultural field delineation task: \texttt{gpt-image-1} via a cloud API, Segment Anything Model~3 (SAM~3) on a remote GPU, and DelineateAnything on a local CPU. The inference backend and protocol are open-source and available at \url{https://github.com/microsoft/nxmndr}.
\end{abstract}

\section{Introduction}

Applying deep learning models to satellite imagery is central to operational workflows across a diverse set of domains, such as agriculture \citep{robinson2026first}, disaster response \citep{krejcar2026ai, akhyar2024deep}, and environmental monitoring \citep{tadesse2024analyzing, sartor2025deep}. However, successful application of these models involves managing model loading, imagery alignment, and result reassembly,  leading to practitioners spending more time writing integration code than running inference. The models that practitioners want to use arrive in heterogeneous formats and target distinct compute environments. Running a single candidate model on a raster covering an area-of-interest requires format-specific deployment code, tile-aligned inference, coordinate transformations, result assembly, and layer registration within geographic information system (GIS) software such as the popular open-source platform QGIS~\citep{QGIS_software}. Multiplied across the number of models a practitioner might evaluate, this integration work, rather than the inference itself, is the dominant bottleneck.

\begin{figure}
    \centering
    \includegraphics[width=0.9\linewidth]{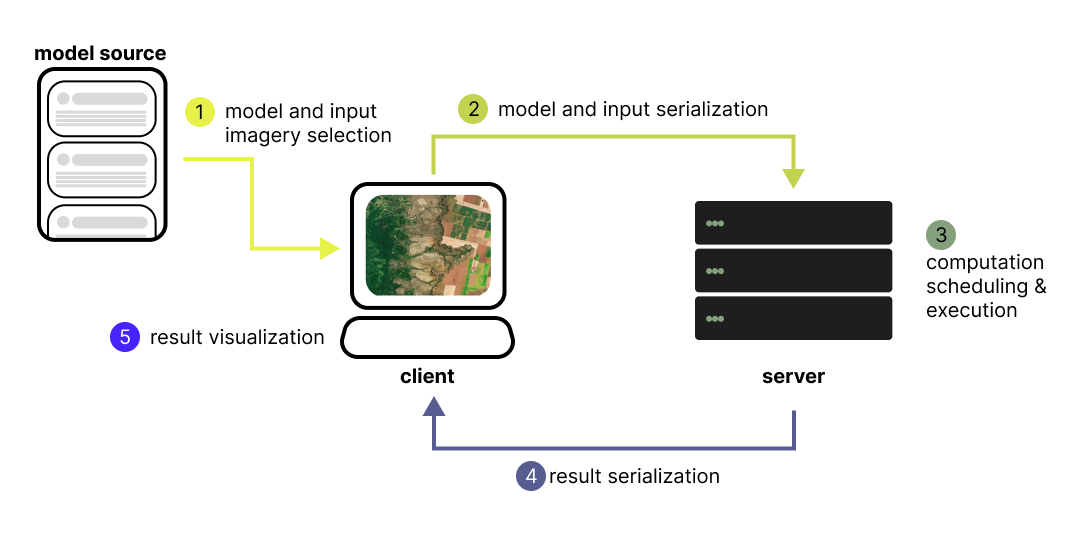}
    \caption{System architecture. (1)~A client selects a model (HuggingFace, cloud API, local weights) and specifies input imagery. (2)~The model specification and tiles stream to the \nxmndr{} server, which dispatches inference to a colocated CPU/GPU; cloud models are reached over HTTP instead, directly or through the \nxmndr{} proxy. (3)~Inference runs on the selected compute target. (4)~Results stream back to the client and (5)~are reassembled, georeferenced, and rendered in the QGIS canvas.}
    \label{fig:architecture}
\end{figure}

To address this gap, we propose Anaximander, a client-server system (Figure~\ref{fig:architecture}) that simplifies model source management and inference execution behind one interactive interface. One protocol separates \emph{where a model comes from} (PyTorch, ONNX, HuggingFace, cloud API) from \emph{where it runs} (local CPU, local GPU, remote GPU, hosted API), so both become a per-run dropdown inside a graphical user interface (GUI) rather than an integration project. The \nxmndr\ backend exposes a remote procedure call (RPC) protocol to load models from PyTorch, ONNX, or HuggingFace sources and run them on a local CPU, local GPU, or remote GPU, plus an HTTP path to cloud application programming interface (API) models. The \nxmndrqgis\ plugin communicates with this backend from inside QGIS, handling tiling, georeferencing, result reassembly, and a live per-tile status overlay, plus a ``Quick Labels'' path that turns an existing QGIS layer legend into a structured prompt for a vision-language model. Spatial navigation inside QGIS becomes the evaluation interface: a practitioner pans or zooms to a region, loads the imagery on which a model should be run, and runs any registered model on the visible extent, observing predictions tile-by-tile as they arrive. The design is extensible: users can register their own model
     sources and compute backends. We release the system as open source to make model evaluations reproducible, to let the community extend the registry of model sources and compute backends, and to lower the barrier for GIS practitioners without machine learning tooling to adopt modern models in their existing workflows. 

Our contributions are:
\begin{itemize}
    \item An inference backend that unifies four commonly-used model source families (PyTorch, ONNX, HuggingFace, cloud APIs) with four compute backends (local CPU, local GPU, remote GPU, serverless cloud) behind one interface: a gRPC protocol serves the PyTorch, ONNX, and HuggingFace models, and cloud APIs are reached over HTTP, directly or through an \nxmndr\ proxy.
    \item A QGIS plugin that turns model-evaluation workflows into spatial-navigation workflows in which practitioners run any registered model on the current map extent without writing code, with live per-tile feedback and automatic georeferencing of results.
    \item An open-source release of the inference backend and protocol under the MIT license at \url{https://github.com/microsoft/nxmndr}, with the release of the plugin pending.
    \item A three-model case study on agricultural field delineation demonstrating code-free comparison across a cloud vision-language model (VLM) API, a remote GPU, and a local CPU.
\end{itemize}

\section{Related Work}
\label{sec:related}

Several QGIS-integrated tools partially address the issues we have identified. Geo-SAM~\citep{zhao2023geosam} delivers interactive segmentation using features derived from the Segment Anything Model (SAM)~\citep{kirillov2023sam} at millisecond latency on a local central processing unit (CPU), but only for the SAM model family. SamGeo~\citep{wu2023samgeo} wraps SAM variants~\citep{carion2025sam3} for geospatial use, runs on either CPU or GPU, and ships a FastAPI server that exposes segmentation over HTTP, so inference need not be colocated with the client. That server is likewise confined to the SAM family and is stateless per request: it accepts one image per call, leaving a client that tiles a raster to issue an independent request per tile. GeoAI~\citep{wu_geoai} is a multi-task, multi-family plugin spanning object detection, semantic and instance segmentation, and vision-language models, loading from PyTorch, ONNX, and HuggingFace sources and running on local CPU, CUDA, or Apple MPS. It also reaches a remotely served model, but only along one path: its vision-language features can target a vLLM endpoint, while its PyTorch, ONNX, and \texttt{timm} models execute in-process. Remote execution is therefore available for one model family rather than across the sources it supports. Deepness~\citep{deepness2022,febrita2025evaluatingaidrivenautomatedmap} supports Open Neural Network Exchange (ONNX) detection, segmentation, and regression with a curated model zoo but is tied to a single model format. IAMAP~\citep{tresson2025iamap} extracts features from a range of foundation models on CPU-class hardware, but is scoped to feature extraction and likewise runs only locally. Each of these tools is constrained on at least one of: the supported model sources, the inference task, or where inference executes. Where remote execution is offered, it is bound to a single model family, and it is exposed as isolated per-request calls rather than as a protocol for streaming a tiled raster. In none of the existing solutions we surveyed is a practitioner able to direct the same inputs to multiple models and choose whether inference occurs, because the inference location is fixed by how the tool was installed and where its server, if any, was started. Additionally, no existing project couples source-agnostic remote delegation to a tile-streaming protocol with server-side session state, which is the combination Anaximander provides.

That each of these solutions has independently re-implemented a serving pathway indicates that a thin GIS integration wrapper over a generic inference server such as Triton~\citep{nvidia_triton}, TorchServe~\citep{pytorch_torchserve}, or Ray Serve~\citep{ray_serve} is insufficient for the specific demands of geospatial ML: model inference is closely tied to session and tile semantics, requires maintaining geographic metadata (coordinate reference system, geotransform, chip grid position), and incurs additional overhead to unify inputs from different imagery sources. Several packages, such as TorchGeo~\citep{stewart2022torchgeo}, handle exactly those operations, but we choose to implement them on top of the GDAL library that ships with QGIS for deployment convenience, so the plugin adds no geospatial dependency to the user's QGIS installation. TorchGeo also bundles a number of model implementations and weights, but is agnostic to where those models are deployed.

\section{Architecture}

Anaximander is composed of two interacting components: the \nxmndr\ server, which loads models and executes inference, and the \nxmndrqgis\ plugin, which drives the server from inside QGIS. The two communicate over a gRPC protocol that any other client (e.g. an interactive notebook, a batch pipeline, or a web application) can reuse in order to provide interactive inference capabilities.

\begin{figure}
    \centering
    \resizebox{\linewidth}{!}{% Figure 2, \input by nxmndr.tex, whose preamble loads TikZ and the libraries used here
% (positioning, arrows.meta, fit, backgrounds, calc).
\begin{tikzpicture}[
    font=\sffamily\small,
    hostbox/.style={
        draw=black, line width=0.5pt, rounded corners=3pt,
        fill=black!4, inner sep=3mm,
    },
    uibox/.style={
        draw=black, line width=0.5pt, rounded corners=2pt,
        minimum height=15mm, text width=38mm,
        align=center, fill=white, inner sep=2mm,
    },
    extbox/.style={
        draw=black, line width=0.5pt, rounded corners=2pt,
        minimum height=12mm, text width=38mm,
        align=center, fill=black!3, inner sep=2mm,
    },
    grpcbox/.style={
        draw=black, line width=0.5pt, rounded corners=2pt,
        minimum height=10mm, text width=126mm,
        align=center, fill=black!8, inner sep=2mm,
    },
    midbox/.style={
        draw=black, line width=0.5pt, rounded corners=2pt,
        minimum height=15mm, text width=38mm,
        align=center, fill=white, inner sep=2mm,
    },
    devicebox/.style={
        draw=black, line width=0.5pt, rounded corners=2pt,
        minimum height=13mm, text width=50mm,
        align=center, fill=black!3, inner sep=2mm,
    },
    arr/.style={-{Stealth[length=2.5mm]}, line width=0.5pt},
    biarr/.style={{Stealth[length=2.5mm]}-{Stealth[length=2.5mm]}, line width=0.5pt},
    groupbox/.style={
        draw=black!55, densely dashed, line width=0.4pt,
        rounded corners=3pt, inner sep=4mm,
    },
    grouplabel/.style={font=\sffamily\scriptsize, inner sep=0pt, anchor=south west},
]

% Columns: the server's three boxes sit at x = -44, 0, 44 mm. Host and server boxes both span
% x = -71..71 mm, and the plugin's two stacks sit one third in from each side of the host.
\coordinate (mid) at (0, 0);
\coordinate (padleft) at (-64mm, 0);
\coordinate (padright) at (64mm, 0);
\coordinate (leftcol) at (-23.67mm, 0);
\coordinate (rightcol) at (23.67mm, 0);

% ---- nxmndr-qgis plugin inside the QGIS host: the tiled path is a vertical stack
\node[uibox] (browser) at (leftcol) {\textbf{Model browser}};
\node[uibox, below=4mm of browser] (config) {\textbf{Inference widget} \\ {\scriptsize task; bands; tiling; status overlay}};
\node[uibox, below=4mm of config] (job) {\textbf{Inference job (QgsTask)} \\ {\scriptsize tile streaming; reassembly; \\[-1.5pt] georeferencing}};
\node[uibox] (quick) at (rightcol |- config) {\textbf{Quick Labels} \\ {\scriptsize legend-driven VLM prompt}};

\node[grouplabel] (pluginlabel) at ([yshift=2mm]browser.north -| padleft) {\textbf{\ttfamily nxmndr-qgis} plugin};
\node[groupbox, fit=(pluginlabel)(padleft)(padright)(browser)(job)(quick)] (plugin) {};

\node[anchor=south, align=center] (hosttitle) at ([yshift=1mm]plugin.north) {\textbf{QGIS host} \\ {\scriptsize map canvas; raster / vector layers; project; renderer}};
\begin{scope}[on background layer]
\node[hostbox, fit=(hosttitle)(plugin)] (host) {};
\end{scope}

\node[extbox, anchor=north] (vlm) at ([yshift=-11mm]host.south -| quick) {\textbf{Cloud VLM API} \\ {\scriptsize Azure AI Foundry / OpenAI}};

% ---- nxmndr server: a local server process or one on a remote host
\node[grpcbox, anchor=north] (grpc) at ([yshift=-16mm]vlm.south -| mid) {\textbf{gRPC service}};

\node[midbox, below=8mm of grpc] (disp) {\textbf{Device dispatcher} \\ {\scriptsize routes each request}};
\node[midbox, left=6mm of disp] (sess) {\textbf{Session manager} \\ {\scriptsize per-session state; TTL cleanup}};
\node[midbox, right=6mm of disp] (mdl) {\textbf{Model registry} \\ {\scriptsize LRU cache}};

\coordinate (spadright) at (67mm, 0 |- grpc.north);
\node[grouplabel, align=left] (serverlabel) at ([xshift=-4mm, yshift=2mm]grpc.north west) {\textbf{\ttfamily nxmndr} server \\ (local server process or remote host)};
\begin{scope}[on background layer]
\node[groupbox, solid, fit=(serverlabel)(spadright)(grpc)(sess)(mdl)] (server) {};
\end{scope}

\node[devicebox, below=10mm of disp] (device) {\textbf{Compute device} \\ {\scriptsize local CPU / local GPU / remote GPU}};

% ---- arrows
\draw[arr] (browser) -- (config);
\draw[arr] (config) -- (job);
\draw[biarr] (job.south) -- node[left, font=\sffamily\scriptsize, xshift=-1mm] {gRPC stream} (grpc.north -| job);
\draw[biarr] (quick.south) -- (vlm.north);
\node[right, align=left, font=\sffamily\scriptsize, inner sep=0pt] at ([xshift=1.5mm]$(host.south -| quick)!0.5!(vlm.north)$) {HTTP};

\draw[arr] (grpc.south -| sess) -- (sess.north);
\draw[arr] (grpc.south -| disp) -- (disp.north);
\draw[arr] (grpc.south -| mdl) -- (mdl.north);

\draw[arr] (disp.south) -- (device.north);

\end{tikzpicture}%
} % inlined TikZ source
    \caption{Anaximander system architecture. Top: the \nxmndrqgis\ plugin integrates into QGIS through the QGIS plugin interface, exposing three user-facing widgets: a Model browser, an Inference widget (task, bands, tiling, live status overlay), and Quick Labels (legend-driven prompts for vision-language models). Tiled inference runs as a background QGIS task (\texttt{QgsTask}) that streams tiles over a single gRPC stream per job to an \nxmndr\ server, either one the plugin starts as a local child process or one on a remote host; Quick Labels bypasses the tiled pipeline, and the plugin's local worker process calls cloud vision-language APIs over HTTP, directly or through the \nxmndr\ proxy. Bottom: in the \nxmndr\ server, a gRPC service drives the inference lifecycle. Incoming requests are mediated by a session manager (per-session state, TTL cleanup), a model registry (LRU cache), and a device dispatcher, which routes each request to a compute device: a local CPU or GPU, or the GPU of the remote host that runs the server.}
    \label{fig:system_arch}
\end{figure}
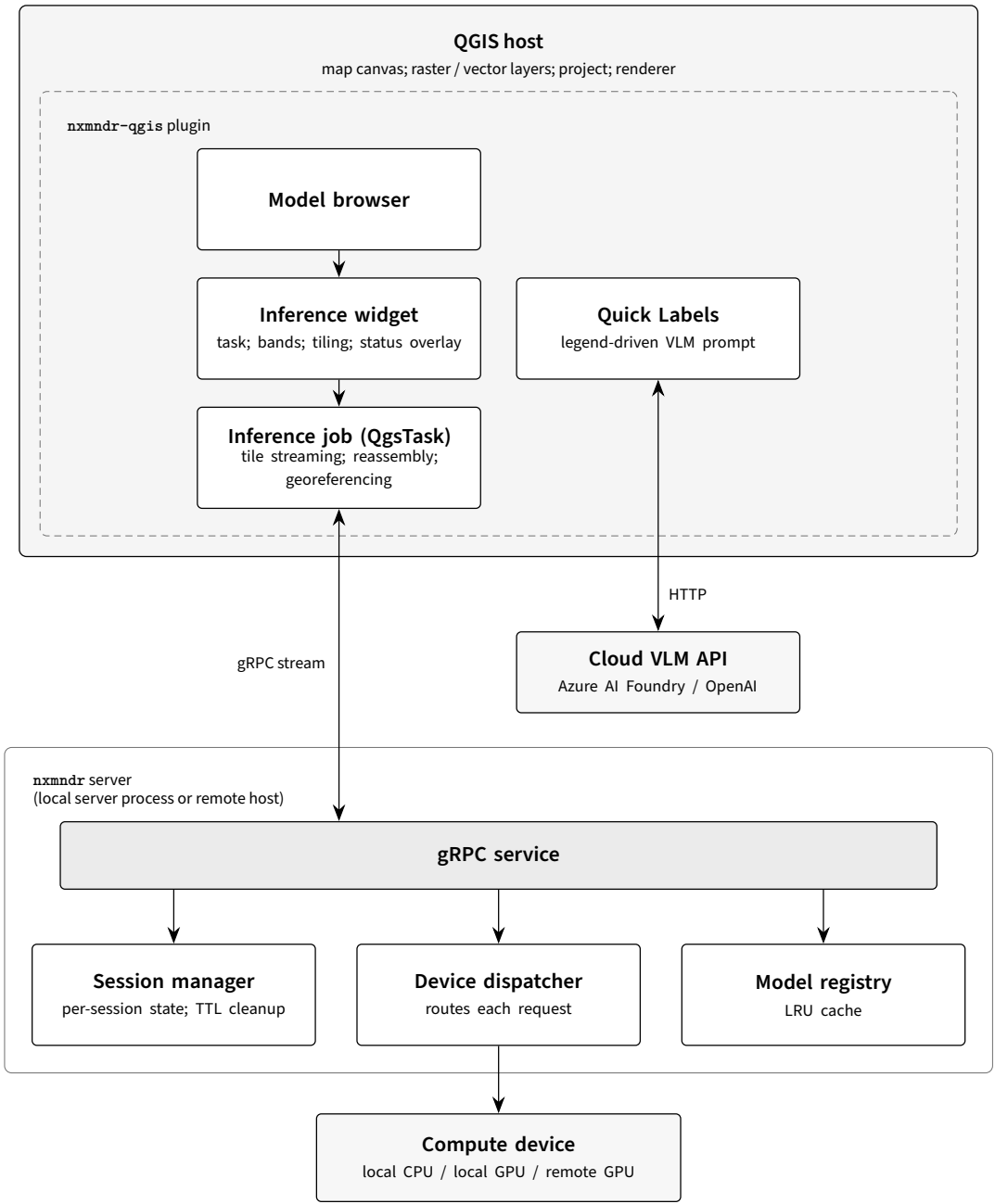

\subsection{Inference Backend (\nxmndr)}

Clients speak to \nxmndr\ through a single service (Figure~\ref{fig:system_arch}) that supports model loading, session management, single-shot prediction, and streaming tiled inference. The core of the service is a bidirectional-streaming channel, in which the client opens one server connection, streams tile chunks, and receives one response per completed tile tagged with the session and tile identifiers, per-tile timing, and the device that served it. This framing collapses thousands of tile submissions into a single long-lived stream, avoiding the connection setup cost that a unary-call-per-tile design would incur.

The models running behind this protocol come from a variety of sources: local PyTorch weights, ONNX graphs, and HuggingFace repositories, and the server unifies them under a shared interface so the rest of the system can treat them identically. The server places every loaded model in a thread-safe LRU cache (default capacity 10) keyed by its specification. For PyTorch and HuggingFace models, the server moves the weights to the chosen device at load time. For ONNX graphs, it creates the ONNX Runtime session with that device's execution provider, so the device is fixed when the session is created rather than chosen per call.

A device dispatcher decides where the request actually runs. At startup the server enumerates the available GPUs (or CPU cores) into a device plan, and a round-robin cursor spreads requests across them. The dispatcher itself is backend-agnostic, and can delegate to specific computation environments. Alongside the gRPC service, the server can run an HTTP proxy to OpenAI or Azure OpenAI that discovers deployments from environment variables; the plugin's Quick Labels path can send its requests through it. Adding a new backend means writing a new provider class, but the remote procedure call surface, session layer, and model registry are unchanged.

Whichever backend runs the inference, a session layer tracks the state that outlives any single RPC call. Each session records the transport capabilities negotiated with the client (chunk size, in-flight tiles, and maximum tile size), the model it serves, per-tile counters for completed and failed tiles, and a creation timestamp. The server answers each tile individually, streaming back one response per tile, and concurrent sessions stay isolated from one another.

\subsection{Model Specification}
Registering new model types with \nxmndr\ is done declaratively: the core library defines a \texttt{ModelSpec} that names the model's \texttt{format} (PyTorch, ONNX, HuggingFace) and its \texttt{source}. Model weight loading and configuration are handled entirely through this interface, so the LRU cache, the session layer, and the device dispatcher are agnostic to the chosen model family, and a new family is added by registering a new spec type.

Each tile response is decoded by result family. A dense output becomes a
single-band label map: leading singleton axes are squeezed, a logit tensor is
arg-maxed over its class axis, a two-dimensional integer map passes through (a
floating-point one is kept as per-pixel values), and a bare class vector becomes
a constant tile. Instance masks, which the model family's adapter on the server
resamples to the chip grid, become an instance map with identifiers unique
within the run. The case study in Section \ref{case_study} exercises the instance path with SAM~3 and DelineateAnything, while the rendered image from \texttt{gpt-image-1} takes the Quick Labels path instead.

\subsection{QGIS Plugin (\nxmndrqgis)}

The plugin frontend (Figure~\ref{fig:system_arch}, top) lets the user select input imagery using standard QGIS interface elements. Custom panels handle communication with cloud-based inference APIs, access to model hubs such as HuggingFace, and selection of local weight files; inference settings (tile size, device location) are specified after model selection.

The plugin hooks into QGIS through its standard plugin-loading mechanism and adds a top-level ``Models'' menu to the QGIS menu bar with three actions: browse models, run Quick Labels, and open settings (Figure~\ref{fig:entry}). Two dock widgets appear on the right-hand side of the main window when the user invokes them (Figure~\ref{fig:panel_location}): a Model Browser for selecting and configuring models, and a Quick Labels dock for legend-driven prompt construction. The plugin attaches all panels, built with QGIS's Qt bindings, through the QGIS plugin interface, so they inherit the host application's raster layers, map extent, and project coordinate reference system without additional plumbing.

To separate machine learning library dependencies from the versions pinned in QGIS's included Python environment, local compute runs in a dedicated worker process with its own Python environment (NumPy~2, PyTorch, ONNX Runtime and the \nxmndr\ package). The plugin starts it as a child process on the first local job or Quick Labels request, and it serves the same \nxmndr\ gRPC service on a loopback port, so a local run uses the same tiled stream as a remote one. The plugin therefore installs no machine learning dependencies into the QGIS environment, and the same worker sends the Quick Labels cloud request.

Once the user selects a model, the Inference Widget replaces the browser (Figure~\ref{fig:inference_settings}). The user picks a task (segmentation, object detection, or embedding generation), maps raster bands to the model's expected R/G/B inputs, sets a chip size and stride (default 512\,pixels, with no overlap), and chooses a compute provider from the list configured in the settings dialog (Figure~\ref{fig:inference_providers}). The tiling widget overlays a live chip grid on the QGIS canvas as rubber-band items and recomputes it whenever the canvas extent changes, so the preview follows the user's pan and zoom.

When the user clicks Run, the Inference Controller queues the run as a QGIS background task, which runs on a QGIS task thread and shows up in QGIS's task manager. The task starts a server session, walks the chip grid and streams requests per tile to the \texttt{nxmndr} server through a client instantiated at the start of the run. As each tile's status changes (pending, in-flight, OK, error, or cancelled) the task emits a signal back to the main thread, and the tile-status overlay recolors in real time (white, yellow, green, red, or dashed gray) as the task continues to dispatch. Because at most a server-negotiated number of tiles are in flight at once, resource use stays predictable; cancelling stops reading new tiles, ends the active stream, sends the server a session-cancel call, and marks every unfinished tile as cancelled.

As tiles complete, the task writes each prediction into its window of a GeoTIFF on disk and flushes it, and the plugin adds that file to the active project as a raster layer as soon as the first tile is written, refreshing the layer as further tiles arrive (at most five times per second) and finalizing its styling when the job ends, so the user watches the result fill in while the server is still responding. The raster is georeferenced on the input layer's own grid (its coordinate reference system and full geotransform) when GDAL can read the layer, or on the canvas coordinate reference system and the rendered extent's grid for a layer it cannot, such as a web tile service, and the user can toggle, style, and overlay it like any other raster. A second, parallel workflow called ``Quick Labels'' (Figure~\ref{fig:gpt-image-1}) bypasses the tiled gRPC pipeline: it extracts class names and RGB values from the renderer of a vector layer the user optionally picks in ``Classes from'' (or uses a default class list), injects them into a prompt template, and sends the prompt together with the current extent's imagery to a cloud vision-language model such as \texttt{gpt-image-1}, hosted on Azure AI Foundry or the OpenAI platform. The plugin georeferences the returned mask and adds it to the project as a raster layer, so the two workflows converge on the same kind of QGIS layer from the user's perspective.

\label{sec:plugin-reassembly}
We handle compositing and seams by allowing the model to see the full tile, but only a centered window of its output is composited into the mosaic. Where
the stride, which sets the size of that window, is smaller than the chip, each tile is therefore predicted with
context on all sides that is then discarded, which suppresses the edge
artifacts that arise when a model is asked to label pixels at the limit of its
receptive field. Partial tiles at the raster margin are padded to a full
chip with the input's nodata value (zero if it has none), marked invalid, and
cropped back after inference, so the grid divides any extent without a
separate boundary case.

Written windows never overlap, because the stride equals the written window, so each output pixel comes from exactly one tile and we perform no blending of class probabilities across tiles. Because compositing is per-pixel, an object crossing a tile boundary is labeled consistently but not \emph{identified} as one object. For field delineation this means a parcel spanning two chips is delineated correctly as area while appearing as two instances, which matters for any downstream count or per-parcel statistic. In line with our philosophy that GIS analysts should be able to use the GUI's capabilities to drive model evaluation and inference, we expect that a user would simply recenter the object split by a seam and run a smaller-scale inference call instead of relying on a general stitching procedure to handle all such tiling artifacts.

In order to keep the client-side memory footprint of the system manageable, input rasters are never loaded entirely into memory. Each chip is a windowed read and each prediction is written to disk as it arrives, so client memory is bounded by chip size and the number of tiles in flight, which the server negotiates, rather than by the extent of the
scene.
A tile whose inference fails is masked as no data in the output raster, so it
stays distinguishable from a background prediction, and it is also surfaced as
an error in the live tile overlay at run time.

\section{Case Study: Agricultural Field Segmentation}
\label{case_study}
Anaximander's core claim is that candidate-model comparison reduces to a sequence of menu selections rather than an integration project. We test this claim on an agricultural field boundary delineation task by running three architecturally different models on identical input imagery, with each model exercising a distinct combination of source family and compute backend: \texttt{gpt-image-1} \citep{openai-gpt-image-1}, a cloud image-generation model driven by a segmentation text prompt, Segment Anything Model~3 (SAM~3)~\citep{carion2025sam3} downloaded from a gated HuggingFace repository onto a remote GPU, and a DelineateAnything \citep{lavreniuk2025delineate} instance loaded from a local weights file on a CPU (Figure~\ref{fig:2x2grid}). The individual outputs illustrate a hypothetical investigation about which model fits this task best, but the primary point of the case study is to demonstrate the user's ability to evaluate all three models without leaving the QGIS interface.

\subsection{Setup}

The input imagery is a 1{,}024\,$\times$\,1{,}024\,pixel crop (physical extent approximately 7.5 km $\times$ 7.5 km) of a Bing Maps satellite scene (\textcopyright~Microsoft) over a well-surveyed \citep{robinson2026first} agricultural region in the Overberg District of Western Cape province in South Africa with visible field boundaries (Figure~\ref{fig:2x2grid}, top-left), loaded as a raster layer in QGIS. All three models receive exactly the same pixels. The plugin and server handle cutting the input into tiles and streaming pixels to, and results from, each model; we keep the plugin's default 512-pixel chip size with no overlap, yielding four tiles per tiled run. The cloud path routes the imagery and a structured instruction prompt (full text in \ref{sec:gpt-prompt}) to OpenAI's \texttt{gpt-image-1} image-editing endpoint and receives a single PNG mask in response. The remote-GPU path fetches SAM~3 weights from HuggingFace Hub, loads them on a GPU node, and ships the four tiles over gRPC. The local-CPU path loads a previously downloaded DelineateAnything weights file from disk into the \nxmndr\ server that the plugin runs in a separate local worker process, and streams the tiles to it over gRPC to run on the host's CPU. We intentionally hold tile size, band mapping (RGB), and input region constant so that any variation in output reflects model behavior rather than configuration.

\subsection{Cloud API: \texttt{gpt-image-1}}
\label{sec:gpt-image-1-desc}

The cloud path begins with \texttt{gpt-image-1} \citep{openai-gpt-image-1}, a multimodal image-generation model from OpenAI that accepts text and image inputs and returns image outputs. Because it is a generation model rather than a native segmentation model, we reach it through the plugin's Quick Labels workflow (a single request to the cloud endpoint for the central 1{,}024\,$\times$\,1{,}024 pixels of the map view) rather than through the tiled inference pipeline. We entered, in the panel's custom-prompt field, a structured instruction prompt asking for a PNG instance-segmentation mask with a unique RGB color per field parcel (full text in \ref{sec:gpt-prompt}). The plugin sends it with the imagery to OpenAI's image-editing endpoint and georeferences the returned PNG into the active QGIS project. Deploying the model required selecting ``Quick Labels'' from the Models menu, choosing the configured \texttt{gpt-image-1} deployment, pasting that prompt into the custom-prompt field, and clicking ``Generate Quick Labels''. The returned mask (Figure~\ref{fig:2x2grid}, top-right) captures the outline of some field parcels but shows substantial boundary inaccuracy, which is consistent with \texttt{gpt-image-1}'s role as a general-purpose image-generation model that treats segmentation as an image-to-image task rather than as dense per-pixel prediction, in line with reported limitations of general-purpose multimodal models on Earth-observation tasks~\citep{zhang2024good}.

\subsection{Remote GPU: Segment Anything Model~3}

The remote-GPU path exercises SAM~3, a promptable foundation segmentation model. We fetched its weights through the plugin's ability to search HuggingFace, deployed them to a remote GPU node, and ran the standard tiled inference pipeline. We ran SAM~3 without geometric or text prompts to keep the comparison between deployment paths fair. Without a concept to score against, SAM~3's per-query confidence is uninformative and standard score thresholding keeps no masks, so we started the server with score thresholding disabled, under which it selects instance masks geometrically instead. Unprompted, SAM~3 produced fragmented masks over roughly a third of the scene, many covering partial parcels rather than whole fields (Figure~\ref{fig:2x2grid}, bottom-left), which is an expected outcome for zero-shot promptable segmentation, not a fair evaluation of SAM~3 itself. The point of this run is that pulling a multi-gigabyte foundation model from a gated HuggingFace repository (requiring authorization handled by plugin configuration), deploying it to a remote GPU, and streaming tiles against it took a similar sequence of clicks as the cloud path.

\subsection{Local CPU: DelineateAnything}

\begin{figure}[H]
    \centering
    \begin{subfigure}{0.45\textwidth}
        \centering
        \includegraphics[width=\linewidth]{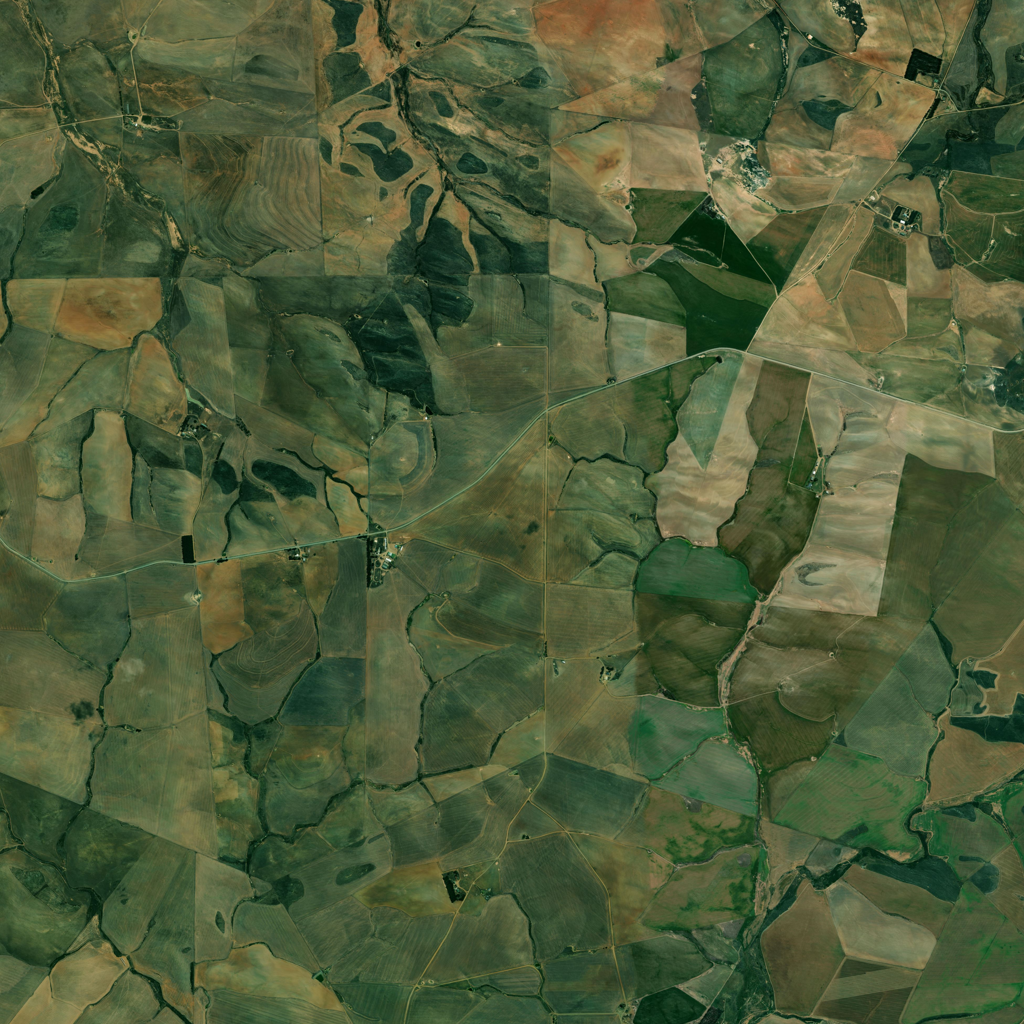}
        \caption{Input imagery (Bing Maps)}
    \end{subfigure}
    \hfill
    \begin{subfigure}{0.45\textwidth}
        \centering
        \includegraphics[width=\linewidth]{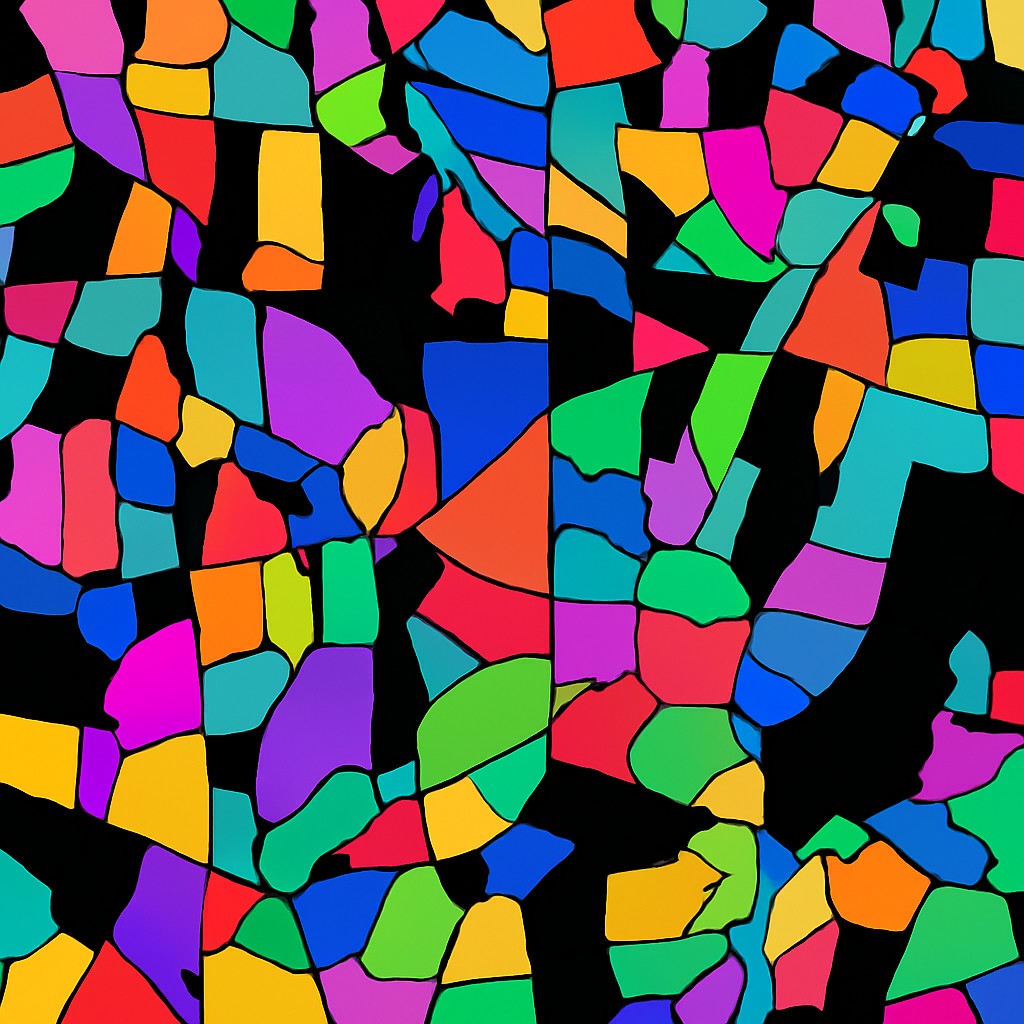}
        \caption{\texttt{gpt-image-1} (cloud API)}
    \end{subfigure}

    \vspace{0.3cm}

    \begin{subfigure}{0.45\textwidth}
        \centering
        \includegraphics[width=\linewidth]{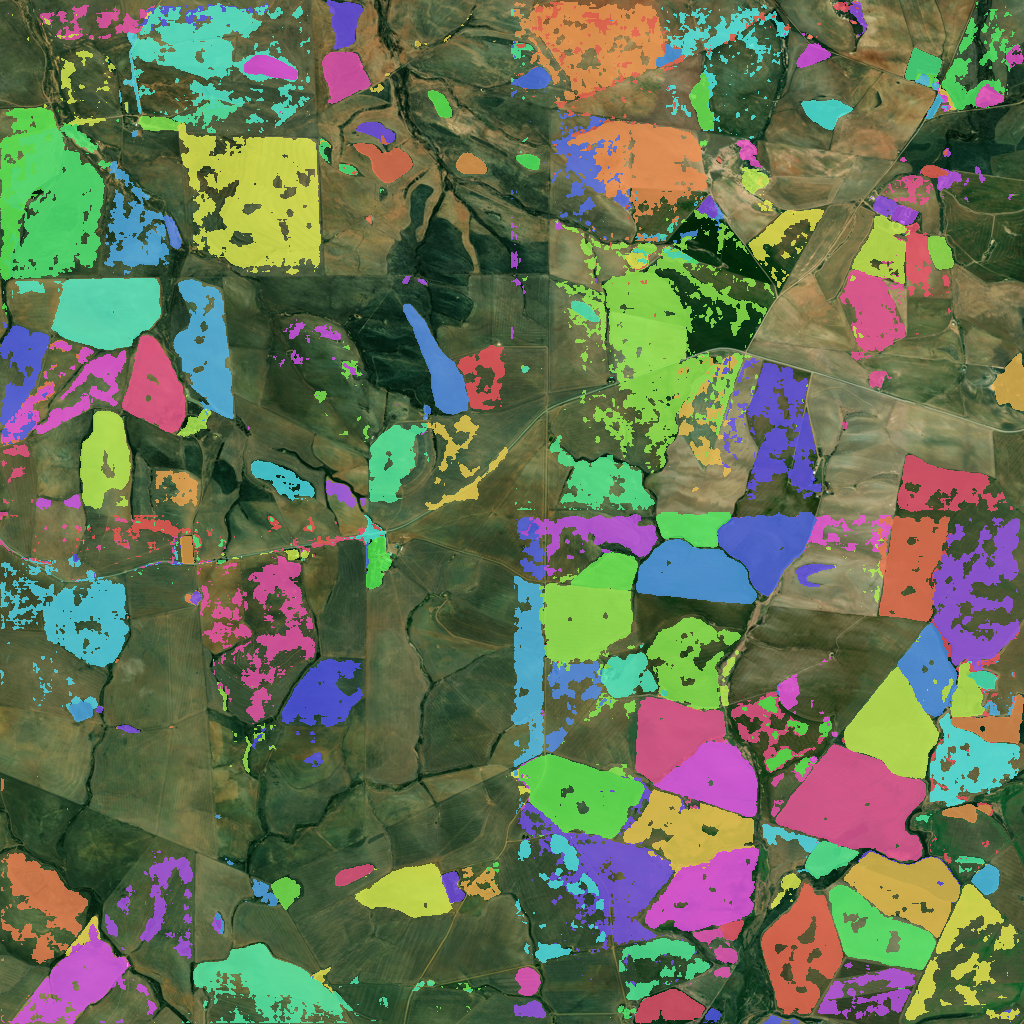}
        \caption{SAM~3 (remote GPU)}
    \end{subfigure}
    \hfill
    \begin{subfigure}{0.45\textwidth}
        \centering
        \includegraphics[width=\linewidth]{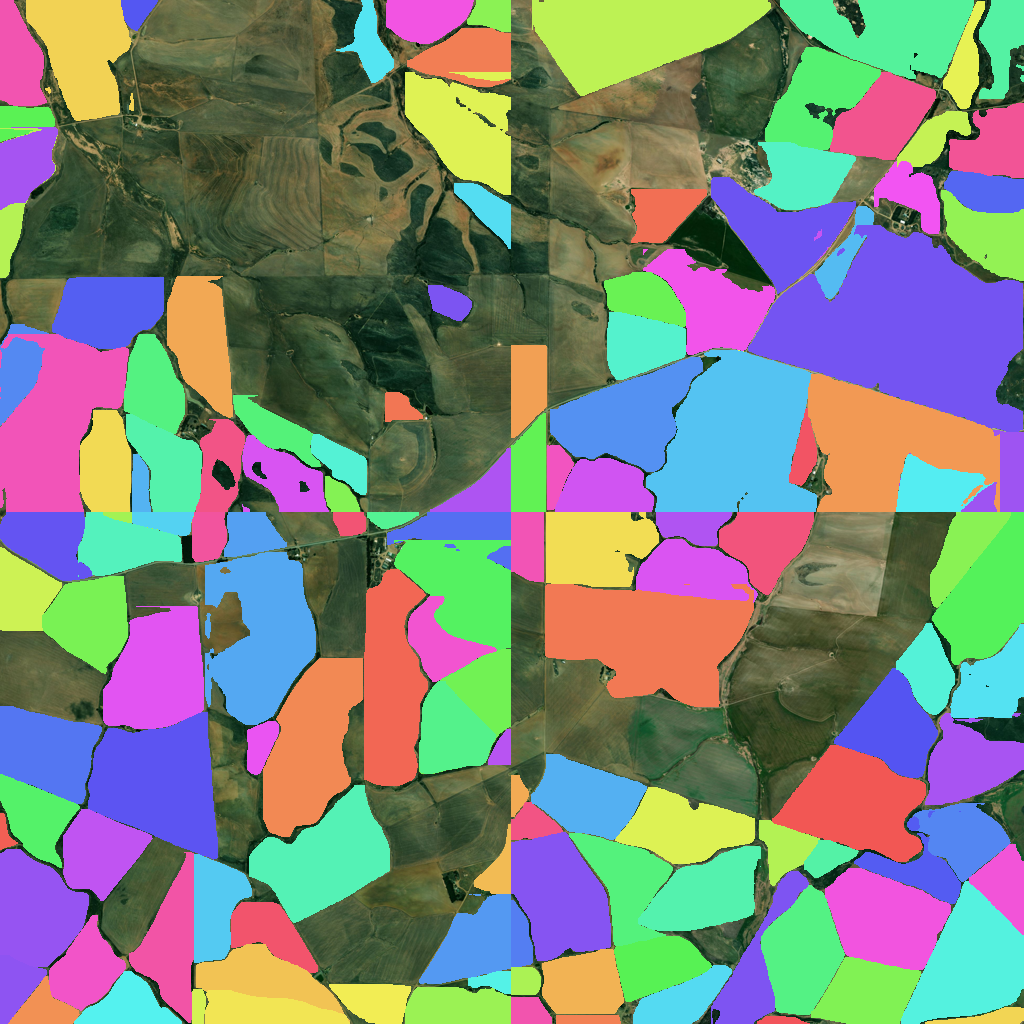}
        \caption{DelineateAnything (local CPU)}
    \end{subfigure}
    \caption{Side-by-side comparison of the input imagery and each model's output, labeled with the deployment path each run exercised. The same plugin UI drove all three runs on identical input imagery; output character is reported qualitatively, not as accuracy: \texttt{gpt-image-1} (OpenAI API, serverless cloud) produced field-like shapes with inaccurate boundaries; SAM~3 (HuggingFace Hub, remote GPU over gRPC) produced fragmented, unprompted masks; and DelineateAnything (local weights file, local CPU) produced tight field boundaries. Input imagery \textcopyright~Microsoft (Bing Maps).}
    \label{fig:2x2grid}
\end{figure}

The local-CPU path exercises DelineateAnything \citep{lavreniuk2025delineate}, a variant of the You Only Look Once (YOLO) detection architecture~\citep{redmon2016you} trained on 22 million field-boundary masks and purpose-built for agricultural field segmentation. We downloaded the weights from the project's GitHub repository\footnote{\url{https://github.com/Lavreniuk/Delineate-Anything/}}, uploaded them through the plugin's local-file picker, selected the local CPU provider from the inference settings, and ran the tiled pipeline with a detection confidence threshold of 0.1. The output (Figure~\ref{fig:2x2grid}, bottom-right) captures most visible field parcels with tight boundaries that closely follow the underlying geometry, with visible artifacts where parcels cross tile edges, which is the expected result for a purpose-built model on in-distribution data run with non-overlapping tiles. This run's purpose is to confirm that the local-CPU path, which shares the tiled streaming protocol with the remote-GPU path, assembles and georeferences CPU-served output correctly, and to provide the qualitative baseline against which the two foundation-model runs are compared.

\subsection{Observations}

This case study is a qualitative demonstration, not a benchmark, and as such, we make no timing, accuracy, or statistical claims. What it does show is the core claim of the paper: the same plugin UI, driven by the same user, ran three models drawn from the source-and-compute matrix Anaximander claims to unify. These models were run on identical imagery, with identical georeferenced raster outputs landing in a single QGIS project. Switching among the three was a menu or dropdown choice, not an integration project: we wrote no Python code, converted no model format, implemented no tiling or georeferencing logic, and imported no per-backend client library. The individual outputs tell a less surprising story (that a purpose-built model fits a narrow task better than two zero-shot foundation models).

\section{Limitations}
In terms of scope, Anaximander's protocol assumes that a model's input is a single-resolution image tensor and that dividing a raster into square tiles covers the map extent. That assumption excludes important classes of geospatial models: those that require multi-scale image pyramids, auxiliary non-image inputs (digital elevation models, temporal stacks, text metadata), or non-image inputs entirely (point clouds, vector features). Supporting these would require extending the server's streaming message schema and the plugin's tile-generation logic. The current design accommodates the dominant pattern in remote-sensing computer vision but does not generalize to the full variety of inputs that recent geospatial foundation models are beginning to expect. Reassembly is similarly simple: each tile writes its own non-overlapping window, with no probability blending across seams, and instance identity is not reconciled across tile boundaries, so an object split by a tile edge appears as two instances (visible as seams in Figure~\ref{fig:2x2grid}, bottom-right). A related scoping choice is that Anaximander focuses exclusively on inference, so users must train or fine-tune models in an external pipeline and export weights in a supported format before deployment, unlike GeoAI or Deepness~\citep{deepness2022}, which bundle lightweight training paths alongside inference. Input preprocessing is future work: the server hands a torchvision or torchgeo model class the chip exactly as sent, unscaled and unnormalized, so pretrained weights from those catalogs need their training-time scaling and normalization exported into the model; a per-model preprocessing field, applied by the server and checked against the catalog's published statistics, is the natural extension.

\section{Conclusion}

Geospatial practitioners have lacked a QGIS-integrated tool that is at once agnostic to where a model comes from and able to orchestrate tiled inference on either local or remote (served) compute through a single streaming protocol. Anaximander addresses this gap: it is an open-source system that unifies four model source families (PyTorch weights, ONNX graphs, HuggingFace repositories, and cloud APIs) and four compute backends (local CPU, local GPU, remote GPU, and cloud) behind a QGIS plugin, with one gRPC protocol for served models and an HTTP path for cloud APIs. Both the model zoos and the compute backend capabilities can be extended by users to fit their needs. Our case study exercises three of those combinations on a single agricultural scene (a cloud vision-language API, a remote GPU running a HuggingFace foundation model, and a local CPU running a YOLO variant) without writing integration code, with georeferenced outputs landing in the same QGIS project in each case.

Several directions extend the system naturally. The most immediate are system benchmarks and a case study on non-RGB imagery. Supporting multi-scale image pyramids, auxiliary non-image inputs, and non-image geometries would broaden the class of models that fit the streaming protocol beyond its current single-resolution-image assumption. Integrating active-learning loops would turn the plugin's tile-status overlay into a human-in-the-loop labeling surface, letting a user correct uncertain tiles without leaving the map. Exposing the server through managed-deployment templates (Kubernetes operators, serverless functions) would simplify large-area mapping for teams without dedicated GPU infrastructure.

A less incremental direction is agentic use. The same gRPC surface that drives the QGIS plugin can drive an LLM-based planner that selects task-specific models dynamically: an autonomous GIS agent \citep{li2023autonomous} can query the \nxmndr\ model registry, pick the appropriate model for each subtask (e.g., building-footprint extraction, crop-type classification, change detection) and stream tiles through it, turning Anaximander into shared infrastructure for interactive and semi-autonomous Earth observation workflows alike.

\printbibliography

\appendix

\section{User Interface Walkthrough}
\label{sec:ui_desc}

This appendix gives a visual tour of the plugin's main workflows. The interface is organized around four stages: locating the plugin inside QGIS, configuring credentials and server endpoints, browsing and selecting a model, and running inference on a raster.

\subsection{Menu entry and panel location}

\begin{figure}[ht]
    \centering
    \begin{subfigure}[b]{0.28\textwidth}
        \centering
        \includegraphics[width=\linewidth]{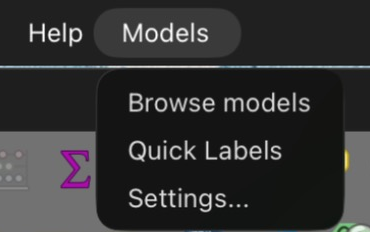}
        \caption{``Models'' menu entry.}
        \label{fig:entry}
    \end{subfigure}
    \hfill
    \begin{subfigure}[b]{0.68\textwidth}
        \centering
        \includegraphics[width=\linewidth]{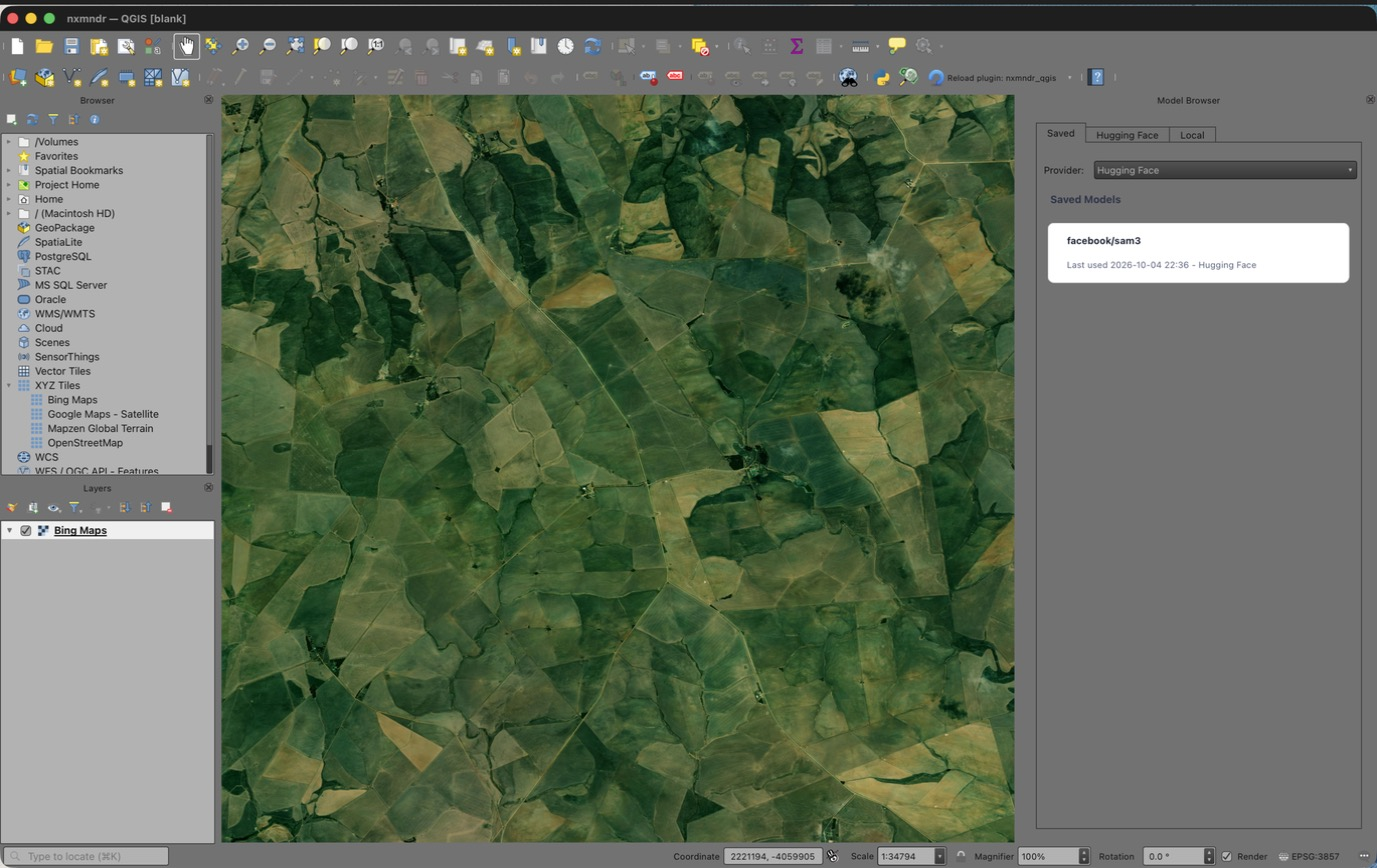}
        \caption{Right-hand panel docked in QGIS.}
        \label{fig:panel_location}
    \end{subfigure}
    \caption{Where the plugin lives: (a) the ``Models'' drop-down in the QGIS menu bar, and (b) the dock panels that appear on the right-hand side of the QGIS main window.}
    \label{fig:entry_and_location}
\end{figure}

The plugin installs a top-level ``Models'' menu in the QGIS menu bar (Figure~\ref{fig:entry_and_location}a). Selecting ``Browse models'' or ``Quick Labels'' from this menu activates a dock widget on the right-hand side of the QGIS main window (Figure~\ref{fig:entry_and_location}b), where all subsequent configuration takes place; the settings action opens a separate dialog.

\subsection{Plugin settings}

\begin{figure}[ht]
    \centering
    \begin{subfigure}[b]{0.65\textwidth}
        \centering
        \includegraphics[width=\textwidth]{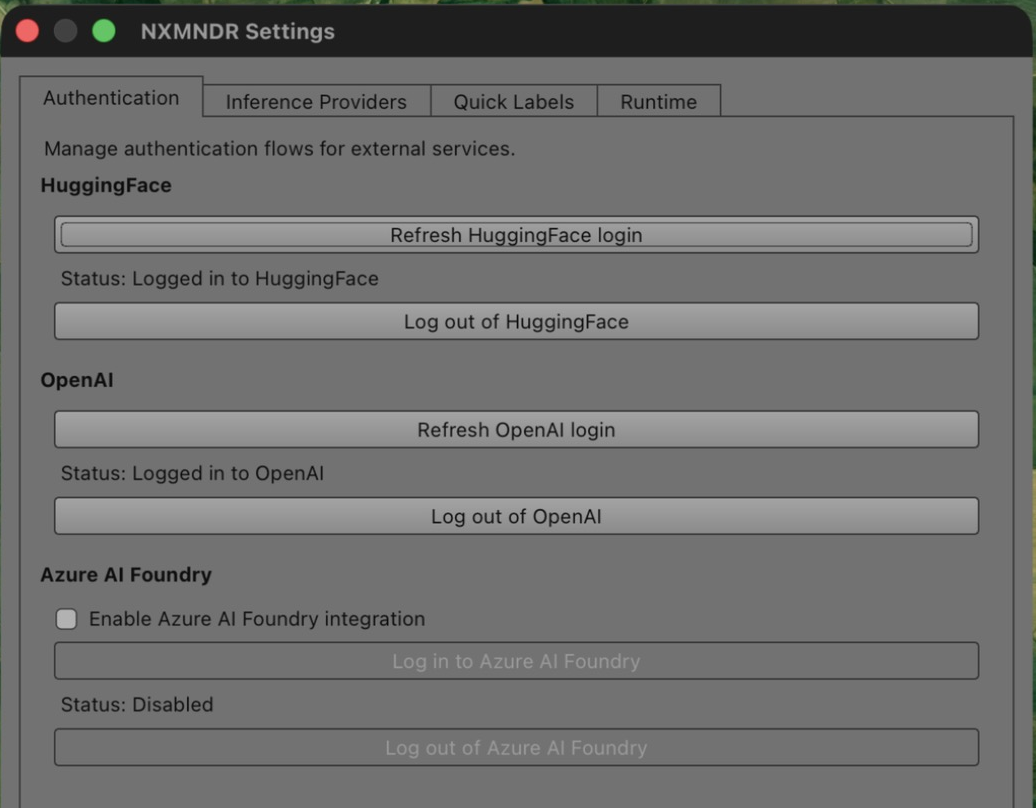}
        \caption{Authentication.}
        \label{fig:auth}
    \end{subfigure}
    \par\medskip
    \begin{subfigure}[b]{0.65\textwidth}
        \centering
        \includegraphics[width=\textwidth]{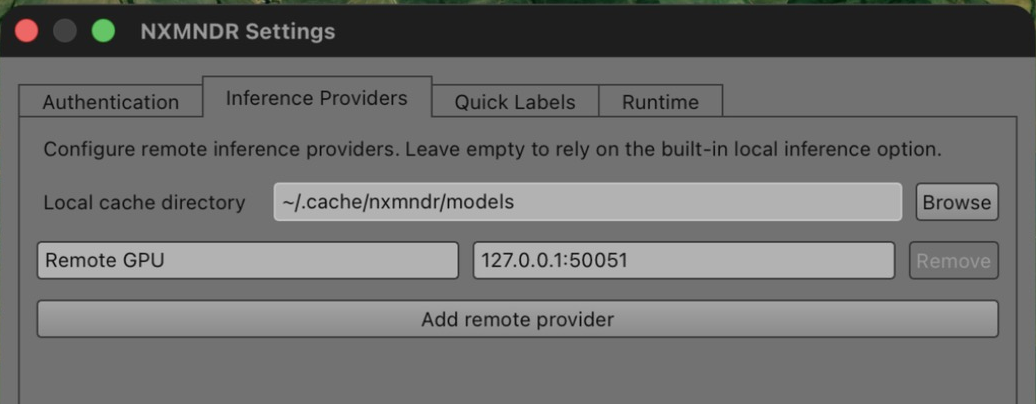}
        \caption{Deployed \nxmndr\ servers.}
        \label{fig:inference_providers}
    \end{subfigure}
    \par\medskip
    \begin{subfigure}[b]{0.65\textwidth}
        \centering
        \includegraphics[width=\textwidth]{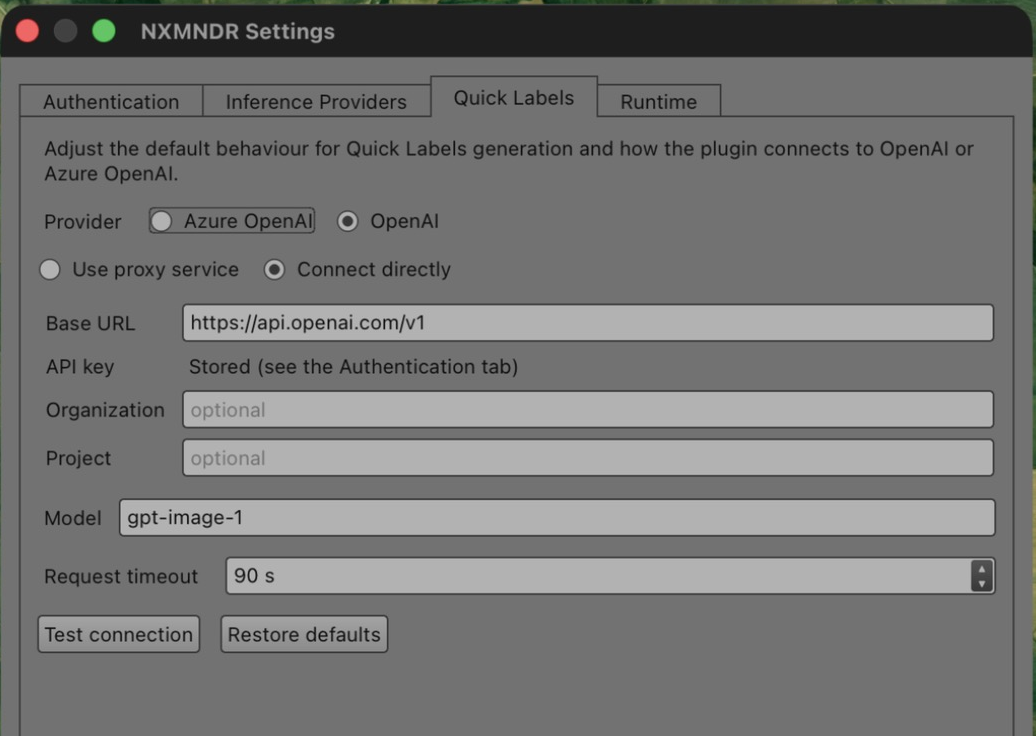}
        \caption{Quick Labels endpoint.}
        \label{fig:azure_ml}
    \end{subfigure}
    \caption{Settings tabs: (a) authentication for services like HuggingFace and Azure AI Foundry, (b) deployed \nxmndr\ server addresses, and (c) the OpenAI or Azure OpenAI endpoint for Quick Labels; the fourth tab, Runtime, is not shown.}
    \label{fig:settings}
\end{figure}

The settings dialog groups four kinds of configuration into separate tabs (Figure~\ref{fig:settings}): sign-ins for HuggingFace, OpenAI and Azure AI Foundry (Authentication), the addresses of deployed \nxmndr\ server instances (Inference Providers), the OpenAI or Azure OpenAI endpoint that Quick Labels calls, directly or through the \nxmndr\ proxy service (Quick Labels), and the environment of the plugin's local worker process (Runtime).

\subsection{Model browser}

Clicking ``Browse models'' surfaces a three-tab panel: Saved, a registry of previously-loaded models regardless of source, capped at twenty entries per inference provider to keep the list browsable; Hugging Face, a HuggingFace Hub search that reuses the user's stored credentials when available; and Local, a local-file picker that accepts a path to weights (filtered to \texttt{.onnx}, \texttt{.pt}, and \texttt{.pth} files). Each tab emits the same selected-model payload, so the downstream inference configuration does not depend on where the model came from. A shared model key derived from the source descriptor (path, repository ID, or endpoint URL) lets the registry collapse duplicate entries for the same model, and the plugin checks at selection time that the server can serve the chosen combination (an ONNX graph, a HuggingFace repository, or a \texttt{.pt}/\texttt{.pth} checkpoint with a model class from the server's list) and refuses any other with a reason; whether the weights load is left to the server, which rejects them when the session opens.

\subsection{Inference configuration}

\begin{figure}[ht]
    \centering
    \begin{subfigure}[b]{0.42\textwidth}
        \centering
        \includegraphics[width=\textwidth]{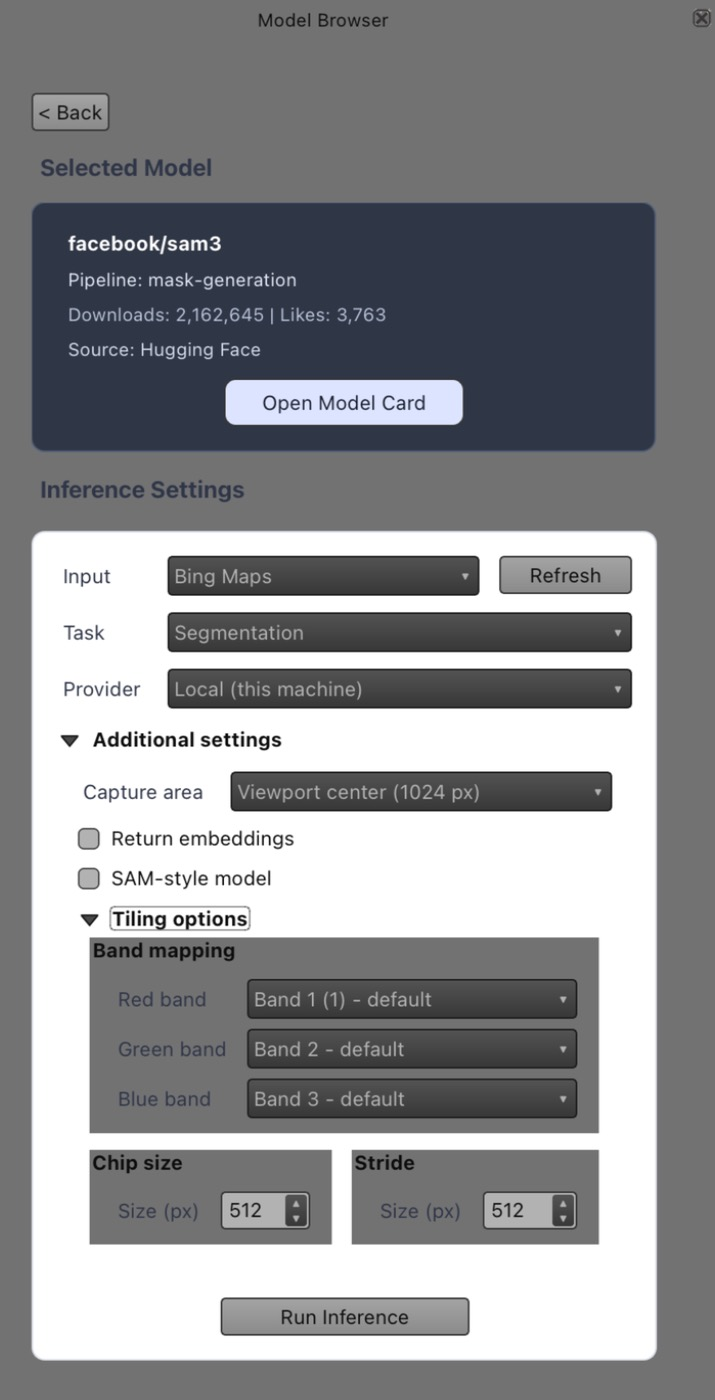}
        \caption{Standard tiled-inference configuration.}
        \label{fig:inference_settings}
    \end{subfigure}
    \hfill
    \begin{subfigure}[b]{0.54\textwidth}
        \centering
        \includegraphics[width=\textwidth]{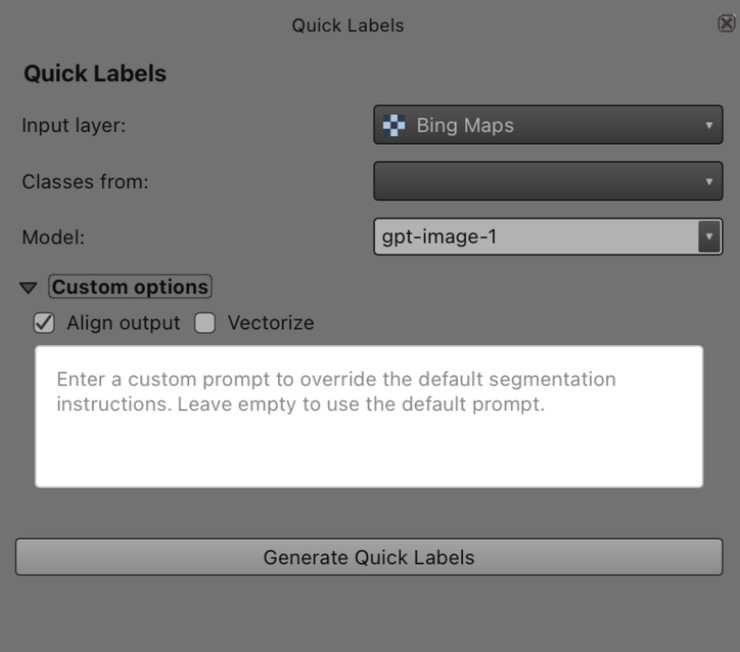}
        \caption{Quick Labels panel.}
        \label{fig:gpt-image-1}
    \end{subfigure}
    \caption{Two inference paths: (a) standard tiled inference through the \nxmndr\ server, and (b) Quick Labels for cloud vision-language models, which bypasses the tiled pipeline.}
    \label{fig:two_workflows}
\end{figure}

After a model is selected, the plugin offers two inference paths (Figure~\ref{fig:two_workflows}). The standard panel drives tiled inference through the \nxmndr\ server: it accepts an input raster layer, a task (segmentation, object detection, or embedding generation), a band mapping, a chip size and stride, and a compute provider. The Quick Labels panel is an alternative path for cloud vision-language models that bypasses the tiled pipeline and calls the cloud endpoint from the plugin's local worker process, directly or through the \nxmndr\ proxy service: it takes an input layer, optional class definitions pulled from an existing layer's legend (which are injected into the prompt), and a configured model endpoint.

\section{Case Study Details}

\subsection{\texttt{gpt-image-1} prompt}
\label{sec:gpt-prompt}

The cloud path in Section~\ref{sec:gpt-image-1-desc} sends \texttt{gpt-image-1} a single composite request combining the input imagery and the following structured instruction prompt:

\noindent
\begin{tcolorbox}[enhanced,breakable,sharp corners,boxrule=0.5pt,arc=0pt,colback=white]
\lstset{basicstyle=\ttfamily\small,breaklines=true,breakatwhitespace=false,columns=fullflexible}
\begin{lstlisting}
You are an instance-segmentation model. Given the satellite image provided, output a PNG instance-segmentation mask with the same resolution as the input, where every distinct agricultural field parcel is detected and assigned its own unique solid RGB color. Each field must be represented as a separate instance, even if adjacent or similar in appearance. Colors must be visually distinct, with no transparency, blending, labels, text, borders, or legend; output only the mask. Each pixel in the image must belong either to one field instance or to a single neutral background color such as black (#000000) that is not reused for any instance. Return only the generated segmentation mask image.
\end{lstlisting}
\end{tcolorbox}
\end{document}